\documentclass{article}
\pdfoutput=1
\usepackage[preprint]{neurips_2023}
\usepackage[utf8]{inputenc} %
\usepackage[T1]{fontenc}    %
\usepackage{xurl}
\usepackage{hyperref}            %
\usepackage{breakurl}          %
\usepackage{amsfonts}       %
\usepackage{amsmath}       %
\usepackage{nicefrac}       %
\usepackage{microtype}      %
\usepackage{lipsum}
\usepackage{graphicx}
\usepackage{makecell}
\usepackage{wrapfig}
\usepackage[size=small]{caption}
\usepackage{mathtools}
\usepackage{amssymb}
\usepackage{amsthm}
\usepackage[algoruled,boxed,lined,noend]{algorithm2e}
\usepackage{adjustbox} 
\usepackage{float}
\usepackage{pifont}
\usepackage{subcaption}
\usepackage{fontawesome}
\usepackage{multirow}
\usepackage{rotating}
\usepackage{array}    
\usepackage{lineno}
\usepackage{longtable}
\usepackage[table,dvipsnames]{xcolor}
\usepackage[numbers]{natbib}
\usepackage{textcomp}
\usepackage{ulem}
\usepackage{color}
\usepackage[most]{tcolorbox}
\usepackage{xcolor}

\definecolor{qual-fig-green}{RGB}{0,144,11}
\definecolor{qual-fig-red}{RGB}{238,0,0}
\definecolor{qual-fig-purple}{RGB}{153,51,255}

\definecolor{answergreen}{RGB}{222,255,222}
\definecolor{promptblue}{RGB}{235,245,255}   
\definecolor{promptgreen}{RGB}{235,255,235}  
\definecolor{promptgray}{RGB}{248,248,248}

\title{Performance of GPT-5 in Brain Tumor MRI Reasoning}

\author{
Mojtaba Safari$^{1}$ \quad Shansong Wang$^{1}$ \quad Mingzhe Hu$^{1}$  \quad Zach Eidex$^{1}$  \quad Qiang Li$^{1}$ \quad \textbf{Xiaofeng Yang}$\textsuperscript{1 \faEnvelope}$ 
\vspace{3mm}
\\
$^1$Department of Radiation Oncology, Winship Cancer Institute, Emory University School of Medicine\\
\faEnvelope \quad Corresponding author: xiaofeng.yang@emory.edu \\
}

\begin{document}
\maketitle

\begin{abstract}
 Accurate differentiation of brain tumor types on magnetic resonance imaging (MRI) is critical for guiding treatment planning in neuro-oncology. Recent advances in large language models (LLMs) have enabled visual question answering (VQA) approaches that integrate image interpretation with natural language reasoning. In this study, we evaluated GPT-4o, GPT-5-nano, GPT-5-mini, and GPT-5 on a curated brain tumor VQA benchmark derived from 3 Brain Tumor Segmentation (BraTS) datasets - glioblastoma (GLI), meningioma (MEN), and brain metastases (MET). Each case included multi-sequence MRI triplanar mosaics and structured clinical features transformed into standardized VQA items. Models were assessed in a zero-shot chain-of-thought setting for accuracy on both visual and reasoning tasks. Results showed that GPT-5-mini achieved the highest macro-average accuracy (44.19\%), followed by GPT-5 (43.71\%), GPT-4o (41.49\%), and GPT-5-nano (35.85\%). Performance varied by tumor subtype, with no single model dominating across all cohorts. These findings suggest that GPT-5 family models can achieve moderate accuracy in structured neuro-oncological VQA tasks, but not at a level acceptable for clinical use.
\end{abstract}


\section{Introduction}
Brain tumors, including glioblastoma multiforme (GBM), meningioma, and brain metastases, represent a diverse group of intracranial neoplasms with distinct radiological appearances, biological behaviors, and prognostic implications~\cite{mcfaline2018brain}. Accurate and timely diagnosis is critical, as early differentiation among these tumor types can significantly influence surgical planning, adjuvant therapy, and patient outcomes~\cite{10.1093/neuonc/noab106}. Magnetic resonance imaging (MRI) remains the clinical standard for brain tumor evaluation due to its superior soft tissue contrast and multiparametric capabilities. However, interpretation of MRI scans requires expert neuroradiological knowledge, and diagnostic variability can arise from differences in training, experience, and fatigue~\cite{laustsen2025interobserver, crowe2017expertise}. As such, the integration of AI, particularly large language and multi-modal models, into the diagnostic workflow has emerged as a promising avenue to enhance efficiency and standardization.

The visual question answering (VQA) paradigm in medical imaging aims to answer clinically relevant, natural-language questions about medical images, combining visual understanding with medical reasoning. In neuro-oncology, VQA systems must not only identify radiological features such as contrast enhancement, necrosis, and peritumoral edema, but also contextualize these findings to provide accurate localization, differential diagnoses, and treatment implications. This makes brain tumor VQA a high-complexity task that demands both robust visual grounding and domain-specific reasoning.

Recent advances in large language models (LLMs) have accelerated a shift away from narrowly trained, task-specific systems toward frameworks in which LLMs function as central reasoning engines~\cite{liu2024deepseek,achiam2023gpt}. In clinical applications, relevant information often arises from diverse modalities. For instance, clinical texts and benchmarks assessing medical knowledge in LLMs~\cite{singhal2023large}, structured representations and grounded reporting frameworks~\cite{guo2023gpt4graphlargelanguagemodels}, and imaging data~\cite{wang2025unifying,wang2025triad}. GPT-4 demonstrated strong general medical reasoning abilities and outperformed prior models across multiple medical benchmarks~\cite{hu2024advancing,  nori2023capabilitiesgpt4medicalchallenge}. However, its performance on brain tumor MRI remained constrained, particularly in fine-grained localization and tumor-type differentiation. The introduction of AutoRG-Brain~\cite{lei2024autorgbraingroundedreportgeneration} highlighted the importance of pixel-level grounding and report generation for brain MRI, providing a structured evaluation framework using BraTS-like datasets that explicitly link visual features to textual outputs. Building on these foundations, GPT-5 and its lightweight variants (GPT-5-mini, GPT-5-nano) have introduced further improvements in multi-modal reasoning~\cite{wang2025capabilitiesgpt5multimodalmedical}~\footnote{\url{https://github.com/wangshansong1/GPT-5-Evaluation}}~\footnote{\url{https://github.com/TsinghuaC3I/MedXpertQA}}, including gains in clinical VQA accuracy, reasoning consistency, and domain adaptability over GPT-4o.

In this study, we evaluated GPT-4o-2024-11-20, GPT-5, GPT-5-mini, and GPT-5-nano on a curated brain tumor VQA benchmark derived from the BraTS datasets. Each case’s multi-modal MRI sequences, T1-contrast (T1c), T1-weighted (T1w), T2-fluid-attenuated inversion recovery (T2-FLAIR), T2-weighted (T2w) are processed to extract centroid-aligned triplanar mosaics. The associated radiology findings are parsed into structured clinical concepts including lesion location, enhancement pattern, boundary definition, edema, midline shift, maximum dimension, and ventricle compression, which are then converted into standardized yes/no and multiple-choice questions with distractors, answer keys, and linked imaging. This ensures that every model is evaluated on identical, clinically grounded prompts, combining both image and text input, and that performance can be directly compared across models and question types. By aligning the dataset construction with~\cite{wang2025capabilitiesgpt5multimodalmedical} prompting and evaluation protocol, we ensure reproducibility and isolate model-specific differences in multi-modal reasoning on glioblastoma, meningioma, and metastasis cases.

\section{Methodology}

This study used the BraTS dataset, which contains multi-parametric MRI scans of glioblastoma (GLI)~\cite{menze2014multimodal}, meningioma (MEN)~\cite{labella2023asnr}, and brain metastases (MET)~\cite{moawad2024brain}. For each case, four standard MRI sequences were available: T1-contrast enhanced (T1c), T1-weighted  (T1w), T2-weighted (T2w), and T2-fluid-attenuated inversion recovery (T2-FLAIR). Tumor segmentation masks provided with the dataset were used to localize lesions, and each case was also paired with radiology-style textual findings generated in the style of AutoRG-Brain or obtained from equivalent clinical reporting sources.

All MRI volumes were converted to a standard RAS orientation prior to analysis to ensure spatial consistency across subjects. Tumor centroids were calculated directly from the segmentation masks. From each volume, axial, sagittal, and coronal slices passing through the tumor centroid were extracted. Intensities were normalized using percentile-based clipping (2nd to 98th percentile) to suppress outliers and enhance lesion visibility. The three slices were concatenated to form a single triplanar mosaic image, providing a compact but information-rich visual representation for subsequent evaluation.

The accompanying clinical text for each case was processed using a custom parsing pipeline to extract structured descriptors of tumor appearance. This process identified the primary lesion location, enhancement pattern, boundary definition, the presence of peritumoral edema, the presence or absence of midline shift, maximum lesion dimension in millimeters, involvement across the midline, and compression of any ventricular structures. The parser combined rule-based term matching, synonym normalization, and regular expressions for numerical values to ensure consistent extraction across reports.

These structured features were then transformed into VQA items following the style protocol described in~\cite{wang2025capabilitiesgpt5multimodalmedical}. Each feature was mapped to a natural-language question designed to assess either visual understanding or clinical reasoning. For example, enhancement features yielded yes/no questions such as ``Does the lesion show ring enhancement?'', location features were converted into multiple-choice questions with plausible distractors, and lesion size measurements were framed as numeric multiple-choice questions. Each VQA item included the question text with labeled answer choices, the correct answer, the associated triplanar mosaic, and metadata specifying the medical task, body system, and question type.

\begin{figure}
	\centering
	\includegraphics[width=\textwidth]{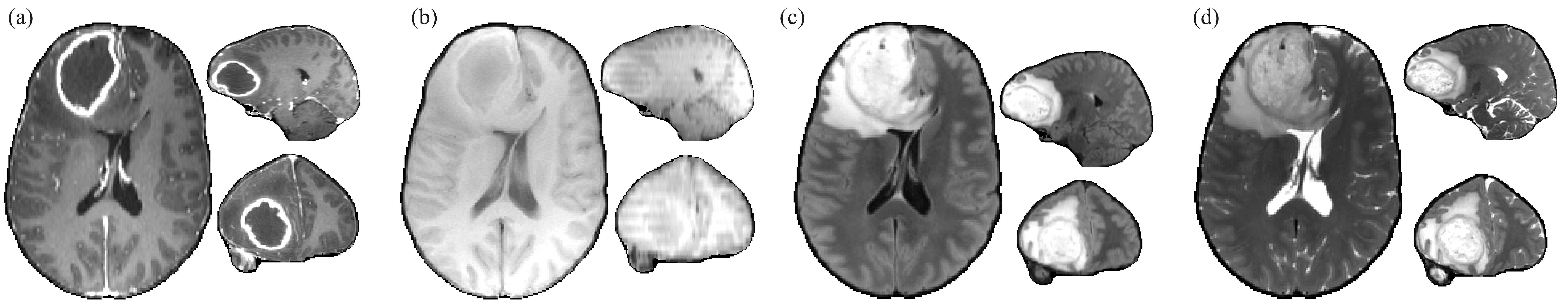}
	\caption{An example BraTS-GLI case containing axial, coronal, and sagittal slices chosen to depict the extent of the tumor volume for (a) T1c, (b) T1w, (c) T2-FLAIR, and (d) T2w MRI. A single sequence was provided to the GPT models for VQA.}
	\label{fig:figure1}
\end{figure}

Four large multi-modal language models were evaluated on the resulting VQA dataset: GPT-4o, GPT-5, GPT-5-mini, and GPT-5-nano. Each model received the triplanar mosaic and the question text with answer choices as a single multi-modal input (see Figure~\ref{fig:figure1}). The model’s predicted choice was compared with the reference answer, and accuracy was calculated as the proportion of correctly answered questions.

\begin{figure}[htbp]
\centering
\begin{tcolorbox}[
        colback=promptgray,
        colframe=black!50!black,
        colbacktitle=white!50!black,
        title=Zero-shot + CoT for VQA,
        fontupper=\normalsize 
    ]
{\linespread{0.8}\selectfont
\begin{verbatim}
[
  {
    "role": "system",
    "content": "You are a helpful medical assistant."
  },
  {
    "role": "user",
    "content": [
      {
        "type": "text",
        "text": "Q: {QUESTION_TEXT}\nA: Let's think step by step."
      },
      {
        "type": "image_url",
        "image_url": { "url": "{IMAGE_URL_1}" }
      }
    ]
  },
  {
    "role": "assistant",
    "content": "{ASSISTANT_RATIONALE}"
  },
  {
    "role": "user",
    "content": "Therefore, among A through {END_LETTER}, the answer is"
  },
  {
    "role": "assistant",
    "content": "{ASSISTANT_FINAL}"
  }
]
\end{verbatim}
}
\end{tcolorbox}

\caption{Prompt design with zero-shot chain-of-thought elicitation. \textcolor{red}{{ASSISTANT\_RATIONALE}} is the model’s intermediate reasoning; \textcolor{red}{{ASSISTANT\_FINAL}} is the final letter used for scoring.}
\label{fig:promptVQA}

\end{figure}

\subsection{Prompt Design}

Each model was presented with a combined input containing the triplanar mosaic image for a single sequence and the associated question text, including the labeled answer choices. Prompts were kept minimal in extraneous instruction to avoid bias from task-specific tuning, while retaining clear, self-contained clinical wording.

In each interaction, the image was provided as the primary visual context, followed immediately by the textual question. Answer choices were explicitly enumerated in the format ``Answer Choices: (A) … (B) … (C) … (D) …'' to standardize parsing across models. Models were explicitly cued with \textcolor{blue}{``Let’s think step by step''} to elicit an intermediate reasoning process (chain-of-thought), captured in \textcolor{red}{ASSISTANT\_RATIONALE}, followed by a final forced-choice selection (\textcolor{red}{ASSISTANT\_FINAL}) consisting of a single letter (A-D). Only \textcolor{red}{ASSISTANT\_FINAL} was used to compute accuracy. Intermediate rationales were logged for qualitative analysis but not graded. This design ensured consistent evaluation while allowing inspection of reasoning behavior.

\begin{figure}[htbp]
\centering
\begin{tcolorbox}[
    colback=promptgray,
    colframe=black!50!black,
    colbacktitle=white!50!black,
    title=A Sample from BraTS-GLI (case BraTS-GLI-00017-000-t1n),
    fontupper=\normalsize,
    enhanced,
    overlay={
        \node[anchor=north east, inner sep=3pt, yshift=-18pt] 
        at (frame.north east) {\includegraphics[width=0.6\textwidth]{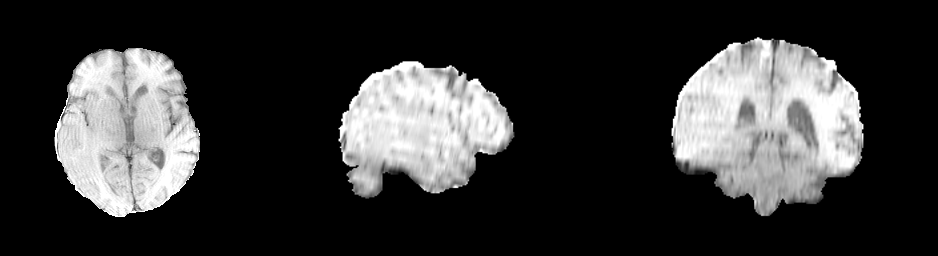}};
    } 
]
\vspace{2cm}
{\linespread{0.8}\selectfont
\begin{verbatim}
[
  {
    "role": "system",
    "content": "You are a helpful medical assistant."
  },
  {
    "role": "user",
    "content": [
      {
        "type": "text",
        "text": "
            Q: Are the lesion boundaries clear or ill-defined         
            Answer Choices: 
                (A) clear               (B) ill-defined 
                (C) mixed/ambiguous       
            A: Let's think step by step."
      },
      {
        "type": "image_url",
        "image_url": { "url": "slices_out_gl/BraTS-GLI-00017-000-t1n_tri.png" }
      }
    ]
  },
  {
    "role": "assistant",
    "content": "{ASSISTANT_RATIONALE}"
  },
  {
    "role": "user",
    "content": "Therefore, among A through C, the answer is"
  },
  {
    "role": "assistant",
    "content": "B"
  }
]
\end{verbatim}
}
\end{tcolorbox}
\caption{Example VQA prompt for a BraTS-GLI case. The triplanar mosaic image is provided first, followed by the question and labeled answer choices. The model outputs an intermediate reasoning step (\texttt{ASSISTANT\_RATIONALE}) and then a final letter (\texttt{ASSISTANT\_FINAL}) corresponding to its chosen answer. Accuracy was computed solely from \texttt{ASSISTANT\_FINAL}.}
\label{fig:promptVQAsample}
\end{figure}

The prompting design template for VQA is shown in Figure~\ref{fig:promptVQA}, and a specific example is shown in Figure~\ref{fig:promptVQAsample}. Prompt wording for each question was taken directly from the feature extraction step without rephrasing or simplification. For example, a detected ring enhancement feature generated the question \textit{``Does the lesion show ring enhancement?''} rather than a paraphrased variant. Location questions retained the anatomical terminology produced by the parsing algorithm, and distractor options were drawn from the same controlled vocabulary used during dataset construction.

By maintaining a fixed input order including image first, then question text, then answer choices, we controlled for positional variation and ensured that all models processed the multi-modal context in a comparable way. The same prompt structure, token limits, and formatting were applied identically to GPT-4o, GPT-5, GPT-5-mini, and GPT-5-nano, allowing performance differences to be attributed to intrinsic model capabilities rather than prompt variability.

\section{Results}
Across the three BraTS cohorts (MET, GLI, MEN), model accuracies clustered in a narrow band, with absolute values ranging from roughly one-third to one-half. Per-cohort results (percent correct) were summarized in Table~\ref{tab:my-table}:

A macro-average (unweighted mean over MET/GLI/MEN) yields: GPT-5-mini 44.19\%, GPT-5 43.71\%, GPT-4o-2024-11-20 41.49\%, and GPT-5-nano 35.85\%. Thus, the highest macro-average was observed for GPT-5-mini, but its advantage over GPT-5 was 0.48 percentage points, which is very small relative to expected sampling variability in VQA settings. Notably, GPT-4o-2024-11-20 achieved the best GLI accuracy (49.80\%), whereas GPT-5 led on MET (42.68\%) and MEN (42.12\%). GPT-5-nano underperformed consistently in all cohorts (see Table~\ref{tab:my-table}).

\begin{table}[]
	\centering
	\caption{Accuracy (\%) across the three BraTS tumor cohorts including brain metastases (MET), glioblastoma (GLI), and meningioma (MEN) and the unweighted macro-average over cohorts. ``$ \Delta $ vs GPT-5'' reports the absolute difference in percentage points relative to GPT-5’s macro-average.}
	\label{tab:my-table}
	\resizebox{0.9\textwidth}{!}{%
		\begin{tabular}{lllll}
			\hline
			Cohort              & GPT-5  & GPT-5-mini & GPT-5-nano & GPT-4o \\\hline
			MET                 & \textbf{42.68} & 42.09      & 35.95      & 38.48             \\
			GLI                 & 46.34 & 48.97      & 38.00         & \textbf{49.80}              \\
			MEN                 & \textbf{42.12} & 41.52      & 33.60       & 36.19             \\
			Average  & 43.71 & \textbf{44.19}      & 35.85      & 41.49             \\
			$ \Delta $ vs GPT-5 & 0.00   & \textbf{0.48}       & -7.86      & -2.22            \\\hline
		\end{tabular}%
	}
\end{table}

\section{Discussion}
The present study evaluated the performance of four LLMs on a curated, clinically grounded brain tumor VQA benchmark derived from BraTS datasets. Across all tumor subtypes, model accuracies clustered within a relatively narrow range, with GPT-5-mini achieving the highest macro-average accuracy (44.19\%), followed closely by GPT-5 (43.71\%) and GPT-4o (41.49\%), while GPT-5-nano consistently underperformed. The marginal differences between GPT-5 and GPT-5-mini suggest that factors beyond model scale may influence performance in this domain.

One potential explanation for GPT-5-mini’s slight advantage is that smaller-scale models may demonstrate a more conservative decision-making process, reducing susceptibility to distractor options in forced-choice formats. Larger models, such as GPT-5, while generally exhibiting stronger general reasoning capabilities, may also be more prone to overinterpret ambiguous features or overfit to irrelevant image-text correlations, particularly in highly structured tasks with minimal prompt context. The variation in performance across tumor types may also reflect the differing radiological characteristics of each cohort. Glioblastomas, with their heterogeneous enhancement, necrosis, and peritumoral edema, may favor models with more generalized reasoning strategies, which could account for GPT-4o’s relative advantage in this category. In contrast, the more stereotyped imaging appearance of meningiomas and metastases may benefit models capable of leveraging structured visual cues more efficiently.

Despite using an identical prompt structure and triplanar mosaic representation across models, multi-modal integration remains a significant challenge. The relatively close clustering of accuracies suggests that, for the present benchmark, no model exhibits a consistently superior visual-textual alignment capability across all neuro-oncological tasks. The forced-choice format further imposes a high cognitive load, requiring precise mapping between subtle visual features and discrete categorical responses, which may limit the performance ceiling across all evaluated systems.

Importantly, our evaluation employed zero-shot chain-of-thought (CoT) prompting, in which models generated an intermediate reasoning step before producing their final answer. While CoT can improve reasoning performance in many general-purpose benchmarks, its impact in this setting is not necessarily positive. In highly structured, forced-choice medical VQA tasks, CoT may encourage overinterpretation of ambiguous or low-signal features, increase susceptibility to distractor options, or reinforce spurious image-text associations. We did not systematically compare CoT with direct-answer prompting, so its net effect on performance in this benchmark remains uncertain. Prior work suggests that in similar settings, CoT may yield only marginal gains or, in some cases, small declines in accuracy due to over-elaboration on irrelevant details~\cite{meincke2025prompting,liu2024mind,jeon2025comparative}.

\subsection{Limitations}

Several limitations should be acknowledged. First, the radiology-style reports used for structured feature extraction were generated by automated pipelines, which may not fully reflect the nuance and variability of natural clinical documentation. Second, the absence of a human expert comparator limits the clinical interpretability of the reported accuracies, as inter-observer variability among neuroradiologists is a relevant point of reference. Third, accuracy was the sole evaluation metric; other important dimensions such as reasoning transparency, uncertainty calibration, and error analysis were not assessed. Fourth, all results reported here were obtained using zero-shot chain-of-thought (CoT) prompting. While this approach can encourage more structured reasoning, it may also lead to overinterpretation of ambiguous features or reliance on spurious image-text associations, potentially reducing accuracy in certain cases. We did not evaluate a direct-answer baseline without CoT, so the net effect of this prompting strategy on performance remains uncertain. Finally, all models were evaluated in a zero-shot setting without domain-specific fine-tuning, which may underestimate the achievable performance in applied clinical scenarios.

\subsection{Future Directions}

Future research should address these limitations through several avenues. Comparative studies involving human experts would clarify the practical utility of these systems in diagnostic workflows. The dataset used in this work already provides paired textual information alongside imaging, which could enable the design of open-form VQA tasks that require models to generate free-text responses. Such an approach would facilitate more nuanced evaluations of GPT models, particularly when benchmarked directly against expert human reasoning. In addition, integrating model calibration metrics and visual grounding evaluations would enhance interpretability and trustworthiness by identifying when models are correct for the wrong reasons or overly confident in incorrect answers. Finally, targeted domain-specific fine-tuning, leveraging neuro-oncology imaging corpora, may further improve performance, particularly for subtle or ambiguous imaging features.

\section{Conclusion}

This study shows that GPT-5 family models achieve moderate accuracy on a standardized, clinically grounded brain-tumor VQA benchmark, with narrow performance differences across variants. These results indicate meaningful, but still limited, capabilities for structured neuro-oncologic reasoning. Importantly, the benchmark itself provides a reproducible, extensible testbed that we expect to be an invaluable resource for future research, enabling systematic comparisons across models, prompts, and training regimes. At the same time, translation to practice will require broader validation: human-expert comparators, evaluations beyond accuracy (e.g., calibration and reasoning transparency), open-form VQA using paired text, and domain-specific fine-tuning across diverse datasets and imaging protocols. Together, these steps will clarify when and how such systems can support real-world clinical workflows.

\bibliographystyle{unsrt}
\bibliography{newbib}

\begin{thebibliography}{10}

\bibitem{mcfaline2018brain}
J~Ricardo McFaline-Figueroa and Eudocia~Q Lee.
\newblock Brain tumors.
\newblock {\em The American journal of medicine}, 131(8):874--882, 2018.

\bibitem{10.1093/neuonc/noab106}
David~N Louis, Arie Perry, Pieter Wesseling, Daniel~J Brat, Ian~A Cree,
  Dominique Figarella-Branger, Cynthia Hawkins, H~K Ng, Stefan~M Pfister, Guido
  Reifenberger, Riccardo Soffietti, Andreas von Deimling, and David~W Ellison.
\newblock The 2021 who classification of tumors of the central nervous system:
  a summary.
\newblock {\em Neuro-Oncology}, 23(8):1231--1251, 06 2021.

\bibitem{laustsen2025interobserver}
Aske~Foldbjerg Laustsen, Rob Dineen, Jurgita Ilginiene, Jonathan~Kj{\ae}r
  Gr{\o}nb{\ae}k, Astrid~Marie Sehested, Kjeld Schmiegelow, Ren{\'e} Mathiasen,
  Marianne Juhler, and Shivaram Avula.
\newblock Interobserver variability in assessing preoperative imaging
  biomarkers for cerebellar mutism syndrome: a multiobserver pilot study.
\newblock {\em Pediatric Radiology}, pages 1--12, 2025.

\bibitem{crowe2017expertise}
Emily~M Crowe, William Alderson, Jonathan Rossiter, and Christopher Kent.
\newblock Expertise affects inter-observer agreement at peripheral locations
  within a brain tumor.
\newblock {\em Frontiers in psychology}, 8:1628, 2017.

\bibitem{liu2024deepseek}
Aixin Liu, Bei Feng, Bing Xue, Bingxuan Wang, Bochao Wu, Chengda Lu, Chenggang
  Zhao, Chengqi Deng, Chenyu Zhang, Chong Ruan, et~al.
\newblock Deepseek-v3 technical report.
\newblock {\em arXiv preprint arXiv:2412.19437}, 2024.

\bibitem{achiam2023gpt}
Josh Achiam, Steven Adler, Sandhini Agarwal, Lama Ahmad, Ilge Akkaya,
  Florencia~Leoni Aleman, Diogo Almeida, Janko Altenschmidt, Sam Altman,
  Shyamal Anadkat, et~al.
\newblock Gpt-4 technical report.
\newblock {\em arXiv preprint arXiv:2303.08774}, 2023.

\bibitem{singhal2023large}
Karan Singhal, Shekoofeh Azizi, Tao Tu, S~Sara Mahdavi, Jason Wei, Hyung~Won
  Chung, Nathan Scales, Ajay Tanwani, Heather Cole-Lewis, Stephen Pfohl, et~al.
\newblock Large language models encode clinical knowledge.
\newblock {\em Nature}, 620(7972):172--180, 2023.

\bibitem{guo2023gpt4graphlargelanguagemodels}
Jiayan Guo, Lun Du, Hengyu Liu, Mengyu Zhou, Xinyi He, and Shi Han.
\newblock Gpt4graph: Can large language models understand graph structured data
  ? an empirical evaluation and benchmarking, 2023.

\bibitem{wang2025unifying}
Shansong Wang, Zhecheng Jin, Mingzhe Hu, Mojtaba Safari, Feng Zhao, Chih-Wei
  Chang, Richard~LJ Qiu, Justin Roper, David~S Yu, and Xiaofeng Yang.
\newblock Unifying biomedical vision-language expertise: Towards a generalist
  foundation model via multi-clip knowledge distillation.
\newblock {\em arXiv preprint arXiv:2506.22567}, 2025.

\bibitem{wang2025triad}
Shansong Wang, Mojtaba Safari, Qiang Li, Chih-Wei Chang, Richard~LJ Qiu, Justin
  Roper, David~S Yu, and Xiaofeng Yang.
\newblock Triad: Vision foundation model for 3d magnetic resonance imaging.
\newblock {\em Research Square}, pages rs--3, 2025.

\bibitem{hu2024advancing}
Mingzhe Hu, Joshua Qian, Shaoyan Pan, Yuheng Li, Richard~LJ Qiu, and Xiaofeng
  Yang.
\newblock Advancing medical imaging with language models: featuring a spotlight
  on chatgpt.
\newblock {\em Physics in Medicine \& Biology}, 69(10):10TR01, 2024.

\bibitem{nori2023capabilitiesgpt4medicalchallenge}
Harsha Nori, Nicholas King, Scott~Mayer McKinney, Dean Carignan, and Eric
  Horvitz.
\newblock Capabilities of gpt-4 on medical challenge problems, 2023.

\bibitem{lei2024autorgbraingroundedreportgeneration}
Jiayu Lei, Xiaoman Zhang, Chaoyi Wu, Lisong Dai, Ya~Zhang, Yanyong Zhang,
  Yanfeng Wang, Weidi Xie, and Yuehua Li.
\newblock Autorg-brain: Grounded report generation for brain mri, 2024.

\bibitem{wang2025capabilitiesgpt5multimodalmedical}
Shansong Wang, Mingzhe Hu, Qiang Li, Mojtaba Safari, and Xiaofeng Yang.
\newblock Capabilities of gpt-5 on multimodal medical reasoning.
\newblock {\em arXiv preprint arXiv:2508.08224}, 2025.

\bibitem{menze2014multimodal}
Bjoern~H Menze, Andras Jakab, Stefan Bauer, Jayashree Kalpathy-Cramer, Keyvan
  Farahani, Justin Kirby, Yuliya Burren, Nicole Porz, Johannes Slotboom, Roland
  Wiest, et~al.
\newblock The multimodal brain tumor image segmentation benchmark (brats).
\newblock {\em IEEE transactions on medical imaging}, 34(10):1993--2024, 2014.

\bibitem{labella2023asnr}
Dominic LaBella, Maruf Adewole, Michelle Alonso-Basanta, Talissa Altes,
  Syed~Muhammad Anwar, Ujjwal Baid, Timothy Bergquist, Radhika Bhalerao, Sully
  Chen, Verena Chung, et~al.
\newblock The asnr-miccai brain tumor segmentation (brats) challenge 2023:
  Intracranial meningioma.
\newblock {\em arXiv preprint arXiv:2305.07642}, 2023.

\bibitem{moawad2024brain}
Ahmed~W Moawad, Anastasia Janas, Ujjwal Baid, Divya Ramakrishnan, Rachit
  Saluja, Nader Ashraf, Nazanin Maleki, Leon Jekel, Nikolay Yordanov, Pascal
  Fehringer, et~al.
\newblock The brain tumor segmentation-metastases (brats-mets) challenge 2023:
  Brain metastasis segmentation on pre-treatment mri.
\newblock {\em ArXiv}, pages arXiv--2306, 2024.

\bibitem{meincke2025prompting}
Lennart Meincke, Ethan Mollick, Lilach Mollick, and Dan Shapiro.
\newblock Prompting science report 2: The decreasing value of chain of thought
  in prompting.
\newblock {\em arXiv preprint arXiv:2506.07142}, 2025.

\bibitem{liu2024mind}
Ryan Liu, Jiayi Geng, Addison~J Wu, Ilia Sucholutsky, Tania Lombrozo, and
  Thomas~L Griffiths.
\newblock Mind your step (by step): Chain-of-thought can reduce performance on
  tasks where thinking makes humans worse.
\newblock {\em arXiv preprint arXiv:2410.21333}, 2024.

\bibitem{jeon2025comparative}
Sohyeon Jeon and Hong-Gee Kim.
\newblock A comparative evaluation of chain-of-thought-based prompt engineering
  techniques for medical question answering.
\newblock {\em Computers in Biology and Medicine}, 196:110614, 2025.

\end{thebibliography}

\end{document}